%% file: main.tex
\documentclass[10pt,twocolumn,letterpaper]{article}
\usepackage[accsupp]{axessibility}
 \usepackage{cvpr}              % To produce the CAMERA-READY version
\input{preamble}

\definecolor{cvprblue}{rgb}{0.21,0.49,0.74}
\usepackage[pagebackref,breaklinks,colorlinks,citecolor=cvprblue]{hyperref}

\def\paperID{31} % *** Enter the Paper ID here
\def\confName{CVPR}
\def\confYear{2024}

\title{\muli{}: Maintaining Unperturbed LiDAR-Event Calibration}

\author{Mathieu Cocheteux$^{1}$, Julien Moreau$^{1}$, Franck Davoine$^2$\\
{ $^1$Université de technologie de Compiègne, CNRS, Heudiasyc, France}\\
{$^2$CNRS, INSA Lyon, UCBL, LIRIS, UMR5205, France}\\
{\tt\small \{mathieu.cocheteux, julien.moreau\}@hds.utc.fr, franck.davoine@cnrs.fr}
}

\begin{document}
\maketitle
\global\csname @topnum\endcsname 0
\global\csname @botnum\endcsname 0
\input{sec/0_abstract}    
\input{sec/1_intro}
\input{sec/2_relatedworks}

\input{sec/3_methodology}
\input{sec/4_experiments}
\input{sec/5_conclusion}

\section*{Acknowledgements}
This work was granted access to the HPC resources on the supercomputer Jean Zay of IDRIS under the allocation 2023-AD011014065 made by GENCI. \\
This work has been carried out within SIVALab, joint laboratory between Renault and Heudiasyc (CNRS / Université de technologie de Compiègne).
{
    \small
    \bibliographystyle{ieeenat_fullname}
    \bibliography{CVPR-WAD}
}


\end{document}

%% file: preamble.tex
\usepackage[dvipsnames]{xcolor}

%% file: sec/0_abstract.tex
\begin{abstract}
Despite the increasing interest in enhancing perception systems for autonomous vehicles, the online calibration between event cameras and \lidar{}—two sensors pivotal in capturing comprehensive environmental information—remains unexplored. We introduce \muli{}, the first online, deep learning-based framework tailored for the extrinsic calibration of event cameras with \lidar{}. This advancement is instrumental for the seamless integration of \lidar{} and event cameras, enabling dynamic, real-time calibration adjustments that are essential for maintaining optimal sensor alignment amidst varying operational conditions. Rigorously evaluated against the real-world scenarios presented in the DSEC dataset, \muli{} not only achieves substantial improvements in calibration accuracy but also sets a new standard for integrating \lidar{} with event cameras in mobile platforms. Our findings reveal the  potential of \muli{} to bolster the safety, reliability, and overall performance of perception systems in autonomous driving, marking a significant step forward in their real-world deployment and effectiveness.
\end{abstract}

%% file: sec/1_intro.tex
\section{Introduction}

\begin{figure}
    \centering
    \includegraphics{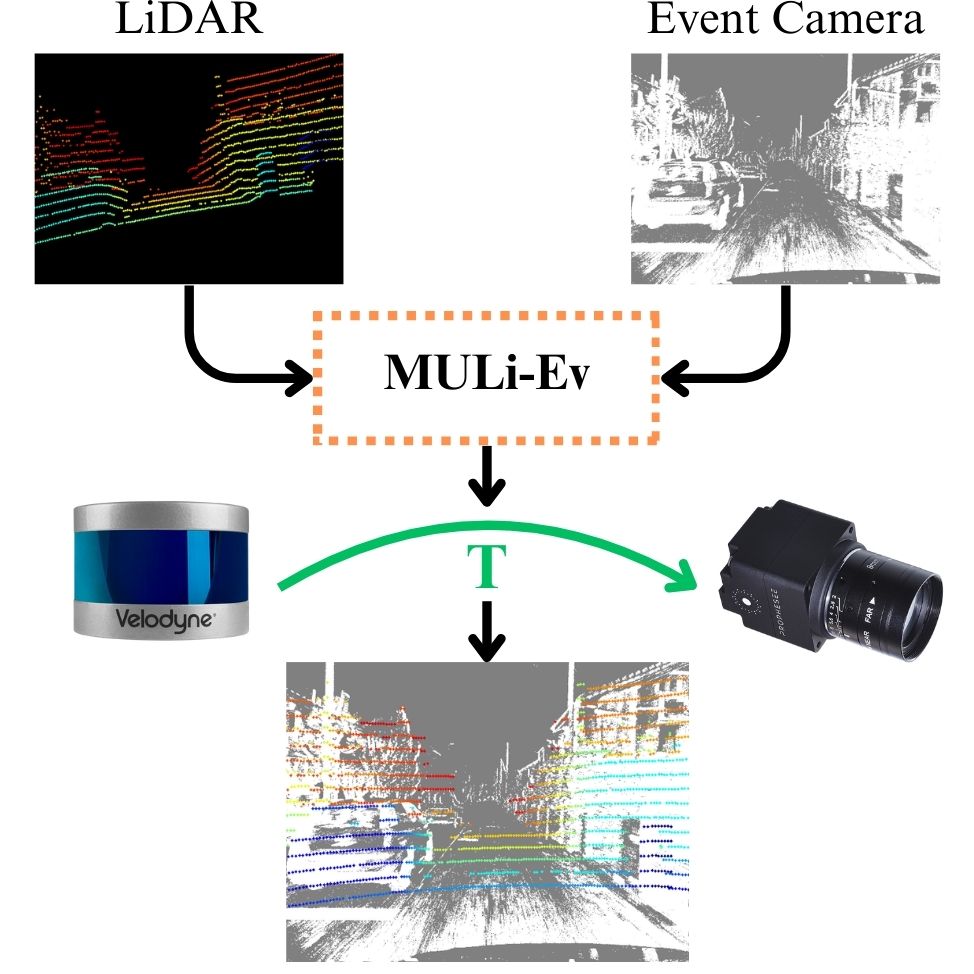}
    \caption{Overview of the \muli{} calibration workflow. This process integrates \lidar{} point clouds and event camera data into the \muli{} network to compute accurate extrinsic calibration parameters (the rigid transformation in $SO(3)$ between the two sensors' reference frames, here represented by T). These parameters enable real-time, precise sensor alignment, facilitating enhanced perception for autonomous vehicles in dynamic scenarios.}
    \label{fig:intro}
\end{figure}

Autonomous driving technologies are on the brink of revolutionizing transportation, announcing a new era of enhanced safety, efficiency, and accessibility. At the heart of this transformation is the development of advanced perception systems that accurately interpret and navigate the complexities of the real world, such as the sharing of the road with other transport modalities (\eg bikes, pedestrians, buses, etc.). A critical element in crafting such systems is sensor calibration. In this work we focus on extrinsic calibration between \lidar{} and event cameras, a subject that still remains too little explored today.

Event cameras, which capture dynamic scenes with high temporal resolution and excel in various lighting conditions, can significantly reduce or help leverage motion blur~\cite{gallego2020event,li2023emergent}. On the other hand, \lidar{} sensors offer detailed depth information vital for precise object detection and environmental mapping. The integration of these complementary technologies promises to substantially elevate vehicle perception capabilities. However, no method has yet been proposed to provide accurate, real-time calibration between these sensors.

Traditional calibration methods~\cite{songCalibrationEventbasedCamera,xingTargetFreeExtrinsicCalibration2023a,ta2023l2e,jiaoLCECalibAutomaticLiDARFrame2023b} perform well under controlled conditions but are unusable in the dynamic, real-world environments autonomous vehicles encounter. These methods often necessitate cumbersome manual adjustments or specific calibration targets, unsuitable for the on-the-fly recalibration needs of operational vehicles. Furthermore, the sparse and asynchronous nature of event camera data introduces additional challenges for the calibration process.

To address these challenges, we propose a novel deep-learning framework trained specifically for the online calibration of event cameras and \lidar{} sensors (\Cref{fig:intro}). This approach not only simplifies the calibration process but also allows onboard online calibration on the vehicle, ensuring consistent sensor alignment. By enabling the joint use of these sensors, our method helps leveraging the complementary strengths of event cameras and \lidar{} in other tasks, significantly enhancing the vehicle's perception system, enabling more accurate object detection and scene interpretation across a diverse range of driving scenarios.

Our contributions include:

\begin{enumerate}
    \item The introduction of a deep-learning framework for online calibration between event cameras and \lidar{}, enabling real-time, accurate sensor alignment—a first for this sensor combination.
    
    \item The validation of our method against the DSEC dataset, showing marked improvements in calibration precision compared to existing methods.
    
    \item The capability for on-the-fly recalibration introduced by our framework directly addresses the challenge of maintaining sensor alignment in dynamic scenarios, a crucial step toward enhancing the robustness of autonomous driving systems in real-world conditions.

\end{enumerate}

The sections that follow will explore related works to contextualize our contributions within the broader research landscape, describe our methodology in detail, present an exhaustive evaluation of our framework against existing state-of-the-art methods, and conclude with a discussion on the broader implications of our findings and potential avenues for future research.

%% file: sec/2_relatedworks.tex
\section{Related Works}

\subsection{Event Camera and \lidar{} Calibration}

The calibration of extrinsic parameters between event cameras and \lidar{} is a necessity to leverage their combined capabilities for enhanced perception in autonomous systems. Unlike traditional cameras, event cameras capture pixel-level changes in light intensity asynchronously, presenting unique challenges for calibration with \lidar{}, which provides sparse spatial depth information. A few offline calibration methods~\cite{songCalibrationEventbasedCamera,xingTargetFreeExtrinsicCalibration2023a,ta2023l2e,jiaoLCECalibAutomaticLiDARFrame2023b} have been proposed.

Song \etal~\cite{songCalibrationEventbasedCamera} made an early contribution with a 3D marker designed for this purpose. Although pioneering, their method necessitates specific, often impractical setup conditions. To address these limitations, Xing \etal ~\cite{xingTargetFreeExtrinsicCalibration2023a} proposed a target-free calibration approach, utilizing natural edge correspondences in the data from both sensors. This innovative method simplifies the calibration process, but is still performed offline. Jiao \etal ~\cite{jiaoLCECalibAutomaticLiDARFrame2023b} introduced {LCE-Calib}, an automatic method that streamlines the calibration process, enhancing robustness and adaptability across various conditions. Building on these advancements, Ta \etal ~\cite{ta2023l2e} introduced L2E, a novel automatic pipeline for direct and temporally-decoupled 6-DoF calibration between event cameras and \lidar{}s, which better leverages the specificities of event data to improve results. 

This progression of techniques underscores a shift towards methods that are not only more versatile but also suited for real-world deployment. However, no method has been proposed until now for the online calibration of this sensor combination.

\subsection{Deep Learning in Extrinsic Calibration}

While deep learning has revolutionized many aspects of autonomous driving technology, its application to extrinsic calibration between event cameras and \lidar{} remains unexplored. Our work introduces the first deep learning-based method for this specific task. However, the groundwork laid by methodologies for RGB cameras and \lidar{} calibration~\cite{Cocheteux_2023_BMVC,cocheteux2023unical,wuThisWaySensors2021,iyerCalibNetGeometricallySupervised2018,lvLCCNetLiDARCamera2021,schneiderRegNetMultimodalSensor2017,jing2022dxq} provides a valuable reference point. For instance, RegNet~\cite{schneiderRegNetMultimodalSensor2017} by Schneider \etal leverages convolutional neural networks (CNNs) for sensor registration, predicting the 6-DOF parameters between RGB cameras and \lidar{} without manual intervention, marking an early milestone in learning-based calibration. Following this, CalibNet~\cite{iyerCalibNetGeometricallySupervised2018} by Iyer \etal further refines the approach with a geometrically supervised network, enhancing the automation and accuracy of the calibration process. 
LCCNet~\cite{lvLCCNetLiDARCamera2021}, introduced by Lv \etal, represents a significant advancement by utilizing a cost volume network to articulate the correlation between RGB images and depth images derived from \lidar{} data, achieving substantial improvements in calibration precision. These methods underscore the potential of integrating deep learning into the calibration workflow, offering insights into feature correlation and end-to-end model training that are instrumental for our approach.

The existing body of work on RGB and \lidar{} calibration delineates a path towards automated, real-time calibration solutions. By adapting and extending these methodologies, our research pioneers the application of deep learning for calibrating event cameras with \lidar{}, aiming to harness the unique advantages of event cameras for enhanced autonomous vehicle perception and navigation.

%% file: sec/3_methodology.tex
\section{Methodology}

Our methodology introduces a deep-learning framework designed for the online calibration of event cameras and \lidar{} sensors, aimed at autonomous driving applications. This section describes the overall architecture of our model, the representation of event data, the calibration process, and details of our training procedure.

\subsection{Architecture}

\begin{figure*}[ht]
\centering
\includegraphics{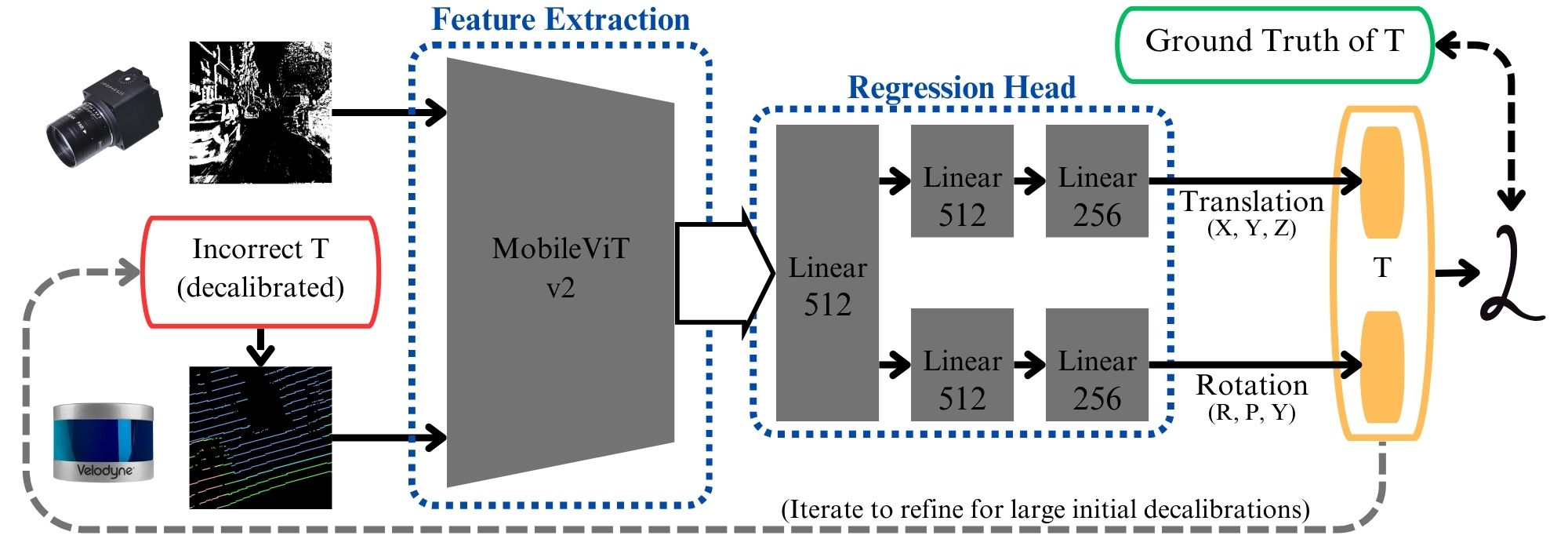}
\caption{Overall architecture of \muli{}. The initial decalibrated extrinsic parameters T (three for rotation, and three for translation) are used to project the \lidar{} point cloud into the event camera frame. Both input are then fed to a MobileViTv2~\cite{mehta2022separable} backbone for feature extraction. The features are passed to a regression head, which regresses separately translation and rotation parameters. Together, they compose the output T, which the loss $\mathcal{L}$ compares to the known ground truth.}
\label{fig:architecture}
\end{figure*}

Our calibration framework integrates event camera and \lidar{} data through a unified deep learning architecture, similarly to UniCal~\cite{cocheteux2023unical}, as illustrated in~\Cref{fig:architecture}. Leveraging a single MobileViTv2~\cite{mehta2022separable} backbone for feature extraction and a custom-designed regression head, the framework achieves precise calibration parameter estimation.

\paragraph{Feature Extraction Backbone:} Central to our approach is the MobileViTv2~\cite{mehta2022separable} backbone, chosen for its fast inference speed and its ability to efficiently process multi-modal data. This facilitates handling event and \lidar{} pseudo-images within a single backbone. By feeding both modalities into separate input channels, our model concurrently processes event camera and \lidar{} data, learning intricate correlations between these two modalities. This unified processing not only streamlines the architecture but also bolsters the model's feature extraction capabilities, crucial for accurate extrinsic calibration.

\paragraph{Custom Regression Head:} Focused on extrinsic calibration parameters, the regression head begins with a common layer that identifies features applicable to both translation and rotation, benefiting from shared data characteristics. Subsequently, the architecture divides into translation and rotation pathways, each comprising two layers designed specifically for their respective parameter sets. This specialization accounts for the unique aspects of translation (x, y, z) and rotation (roll, pitch, yaw) parameters, such as scale and unit differences, thereby enhancing the model's calibration precision.

\subsection{Event Representation}
\label{sec:representations}
\begin{table}
    \centering
    \resizebox{\columnwidth}{!}{%
    \begin{tabular}{@{}lccc@{}}
    \toprule
    \textbf{Event Representation} & \textbf{Dimensions} & \textbf{Polarity} & \textbf{Temporality} \\
    \midrule
    Event Frame & \(H \times W\) & \xmark & \xmark \\
    Voxel Grid & \(B \times H \times W\) & \xmark & \cmark \\
    Time Surface & \(H \times W\) & (\cmark) & (\cmark) \\
    \bottomrule
    \end{tabular}
    }
        \caption{Comparison of some event representations considered for our method, and their properties. H and W represent respectively height and width.}
        \label{tab:event-rep-comparison}
\end{table}

In developing our calibration framework, a critical consideration was the optimal representation of event data captured by event cameras. Event cameras generate data in a fundamentally different manner from traditional cameras, recording changes in intensity for each pixel asynchronously. The data is structured as a flow of events, necessitating a thoughtful binning approach to transform it into a new representation for effective processing and integration with \lidar{} data.

Our investigation encompassed various formats for representing event data, including:
\begin{itemize}
    \item The event frame~\cite{rebecq2017real} representation, which accumulates events into a 2D image, where the intensity of a pixel corresponds to the number of events that occurred at that location within the specified accumulation time. 
    \item The voxel grid~\cite{zhu2019unsupervised} representation, which extends this concept into three dimensions, adding a temporal depth to the accumulation.
    \item The time surface~\cite{benosman2013event} representation, which encodes the most recent timestamp of an event at each pixel, capturing the temporal dynamics more explicitly.
     
\end{itemize}

Each binning strategy offers distinct advantages in terms of capturing the spatial and temporal dynamics of the scene, which are recapitulated in~\Cref{tab:event-rep-comparison}.
However, our primary objective was to identify a representation that not only simplifies the calibration process but also enhances performance by preserving essential geometric information such as edges, without unnecessarily complicating the model with temporal details that are less critical for our specific calibration task.

Ultimately, we found that event frame representation was the most effective approach. This decision was driven by several key factors:
\begin{itemize}
    \item \textbf{Simplicity}: The event frame representation aligns closely with conventional data types used in deep learning, allowing for a more straightforward integration into our calibration framework.
    \item \textbf{Performance}: Through empirical testing (detailed in~\Cref{sec:ablation}), we observed that the event frame provided superior performance in terms of calibration accuracy. This improvement is attributed to the format's effectiveness in preserving the geometric integrity of the scene.
\end{itemize}

In summary, the event frame representation emerged as the superior choice for our online calibration method, balancing simplicity, performance, and geometric fidelity. This finding underscores the importance of matching the data representation format with the specific requirements of the task, especially in the context of sensor fusion and calibration.

\subsection{Training Procedure}

\subsubsection{Model Training}
We introduce artificial decalibrations into the dataset, akin to the strategy employed by RegNet~\cite{schneiderRegNetMultimodalSensor2017}. This involves systematically applying random offsets to the calibration parameters between the event cameras and \lidar{}. The network is then tasked with predicting these offsets, effectively learning to correct the artificially induced decalibrations. The smallest range used during our training focuses on recalibrating the most common yet most challenging and subtle decalibrations within $\pm1^{\circ}$ and $\pm10cm$. However, our model is capable, using an approach similar to~\cite{schneiderRegNetMultimodalSensor2017,lvLCCNetLiDARCamera2021,Cocheteux_2023_BMVC,iyerCalibNetGeometricallySupervised2018}, to correct larger decalibrations, by iterating through a cascade of networks trained on larger decalibrations. For our experiments, we use a cascade of two networks. A first network with a larger training range of up to $\pm10^{\circ}$ and $\pm100cm$, giving us a rough estimate of the parameters (with an average error of $0.47^{\circ}$ and $3.03cm$), well within the training range of the second network, trained on the $\pm1^{\circ}$ and $\pm10cm$ range.

\subsubsection{Optimization and Evaluation} Throughout the training, we employ Mean Square Error (MSE) regression losses (wildly used for regression tasks, and specifically on calibration tasks~\cite{schneiderRegNetMultimodalSensor2017}) to minimize the difference between the predicted calibration parameters and the ground truth, derived from the original, unaltered DSEC~\cite{gehrig2021dsec} data. The model is trained with the Adam optimizer and a learning rate of $0.0001$. Continuous evaluation on a validation set, separate from the training data, allows us to monitor the model's performance and adjust the training parameters accordingly to avoid overfitting and ensure optimal generalization.

%% file: sec/4_experiments.tex
\section{Experiments}

The effectiveness of our proposed deep-learning framework for online calibration of event cameras and \lidar{} sensors is demonstrated through a series of experiments using the DSEC \cite{gehrig2021dsec} dataset. This section outlines our experimental setup, evaluation metrics,  comparisons with existing works, and the results achieved.

\subsection{Dataset}
\label{sec:dataset}
For our experiments, we leverage the DSEC dataset ~\cite{gehrig2021dsec}, a pioneering resource offering high-resolution stereo event camera data for driving scenarios and \lidar{}. More specifically it relies on a Velodyne VLP-16 \lidar{} (a 16 channels LiDAR), and Prophesee Gen3.1 monochrome event cameras with a $640 \times 480$ resolution. This dataset is particularly notable for its inclusion of challenging illumination conditions, ranging from night driving to direct sunlight scenarios, as well as urban, suburban, and rural environments, making it an ideal benchmark for our calibration framework. Its composition is detailed in~\Cref{tab:dataset-summary}.

\begin{table}
    \centering
    
    \resizebox{\columnwidth}{!}{%
    \begin{tabular}{@{}llllc@{}}
    \toprule
        \textbf{Split} & \textbf{Area} & \textbf{Time} & \textbf{Environment} & \textbf{Sequences} \\
    \midrule
        Training & Interlaken & Day & Rural & 5 \\
        & Thun & Day & Suburban & 1 \\
        & Zurich City & Day/Night & Urban & 35 \\
        Test & Interlaken & Day & Rural & 3 \\
        & Thun & Day & Suburban & 2 \\
        & Zurich City & Day/Night & Urban & 7 \\
    \bottomrule
    \end{tabular}

    }
    \caption{Subsets of the DSEC~\cite{gehrig2021dsec} dataset by location of capture, and their characteristics.}
    \label{tab:dataset-summary}
\end{table}

\subsection{Preprocessing}

Preprocessing temporally aligns \lidar{} and event camera data for our calibration framework, before entering the network. The steps include:

\begin{itemize}
    \item \textbf{Projection of \lidar{} Data:} Initial (erroneous) calibration parameters are used to project \lidar{} point clouds into the event camera frame.
    \item \textbf{Temporal Synchronization:} \lidar{} timestamps are used to synchronize the \lidar{} data with asynchronous events from the event camera, ensuring accurate event accumulation over \lidar{} scans.
    \item \textbf{Event Accumulation:} A 50ms window (evaluated in~\Cref{sec:ablation}) is used for event accumulation, balancing scene representation detail with data volume.
    \item \textbf{Transformation to Event Frame:} Accumulated events are converted into an event frame (also evaluated in~\Cref{sec:ablation}), preparing the data for neural network processing.
\end{itemize}

Data normalization is applied as a standard step, bringing \lidar{} and event camera data to a same scale for optimal feature extraction.

\subsection{Evaluation Metrics}
We employ the Mean Absolute Error (MAE) to gauge the accuracy of our calibration technique, both for translational and rotational parameters. The MAE for translation components is defined as the average of the absolute discrepancies between the predicted and actual translation vectors, with each component's error given by:
\begin{equation}
    \text{MAE}_{\text{trans}} = \frac{1}{N} \sum_{i=1}^{N} \left\lVert \mathbf{t}_{\text{pred},i} - \mathbf{t}_{\text{gt},i} \right\rVert_2,
\end{equation}
where \( \mathbf{t}_{\text{pred},i} \) and \( \mathbf{t}_{\text{gt},i} \) represent the predicted and ground truth translation vectors for the \(i\)-th sample, respectively, and \( N \) denotes the number of test samples.

For rotation, our network outputs Euler angles, which are converted into rotation matrices to facilitate a robust error computation. The angles are then converted back into Euler form to report errors in a more interpretable fashion. Consequently, the MAE for rotational components—Roll, Pitch, and Yaw—is calculated as follows:
\begin{equation}
    \text{MAE}_{\text{rot}} = \frac{1}{N} \sum_{i=1}^{N} \left\lVert \text{Euler}(\mathbf{R}_{\text{rel},i}) \right\rVert,
\end{equation}
where \( \mathbf{R}_{\text{rel},i} \) represents the relative rotation matrix for the \(i\)-th sample, obtained by the operation \(\mathbf{R}_{\text{pred},i} \times \mathbf{R}_{\text{gt},i}^{-1}\). The function \( \text{Euler}(\cdot) \) converts this matrix to Euler angles, expressing the rotational discrepancy in terms of Roll, Pitch, and Yaw. The norm \( \left\lVert \cdot \right\rVert \) then quantifies the magnitude of these angles, yielding the rotational error in degrees. This methodology allows for a precise measurement of rotational calibration performance across the dataset.

\subsection{Experimental Results}
\begin{table}
    \centering
    \resizebox{\columnwidth}{!}{%
    \begin{tabular}{lcccc}
    \toprule
        \textbf{Method} & \begin{tabular}[c]{@{}c@{}}\textbf{Translation} \\ \textbf{Error (cm)}\end{tabular} & \begin{tabular}[c]{@{}c@{}}\textbf{Rotation} \\ \textbf{Error (deg)}\end{tabular} & \textbf{Online} & \begin{tabular}[c]{@{}c@{}}\textbf{Execution} \\ \textbf{Time (s)}\end{tabular} \\ \midrule
        L2E~\cite{ta2023l2e} & \NA & \NA& No & 134 \\ 
        LCE-Calib~\cite{jiaoLCECalibAutomaticLiDARFrame2023b} & 1.5 & 0.3 & No & \NA \\ 
        \muli{} (Ours) & \textbf{0.81} & \textbf{0.10} & \textbf{Yes} & \textbf{$<$~0.1}\\ \bottomrule
    \end{tabular}%
    }
    \caption{Comparison of~\muli{} to the state of the art.}
    \label{tab:calibration_comparison}
\end{table}

Existing methods~\cite{songCalibrationEventbasedCamera,xingTargetFreeExtrinsicCalibration2023a,jiaoLCECalibAutomaticLiDARFrame2023b, ta2023l2e} being offline approaches, their authors chose to evaluate them on a few scenes that they captured themselves. However, considering our online, deep learning-based approach, we evaluated our method in a more systematic way, on the publicly available DSEC~\cite{gehrig2021dsec} dataset presented in~\Cref{sec:dataset}.
Moreover, most existing works measure the quality of their results through non-absolute, sensor-dependent metrics, such as reprojection error, which is more suitable when using targets, and can be affected by sensor resolution and lens distortion. 
One of the most recent works, LCE~\cite{jiaoLCECalibAutomaticLiDARFrame2023b}, is the most suitable for comparison with our method, as it not only offers state-of-the-art results, but also uses similar sensors (notably the same LiDAR, Velodyne VLP-16). It also communicates results in the same absolute metric as our work, measuring the Mean Absolute Error on rotation and translation.

\begin{figure*}
  \centering
  \begin{subfigure}{.5\textwidth}
    \centering
    \includegraphics[width=\linewidth]{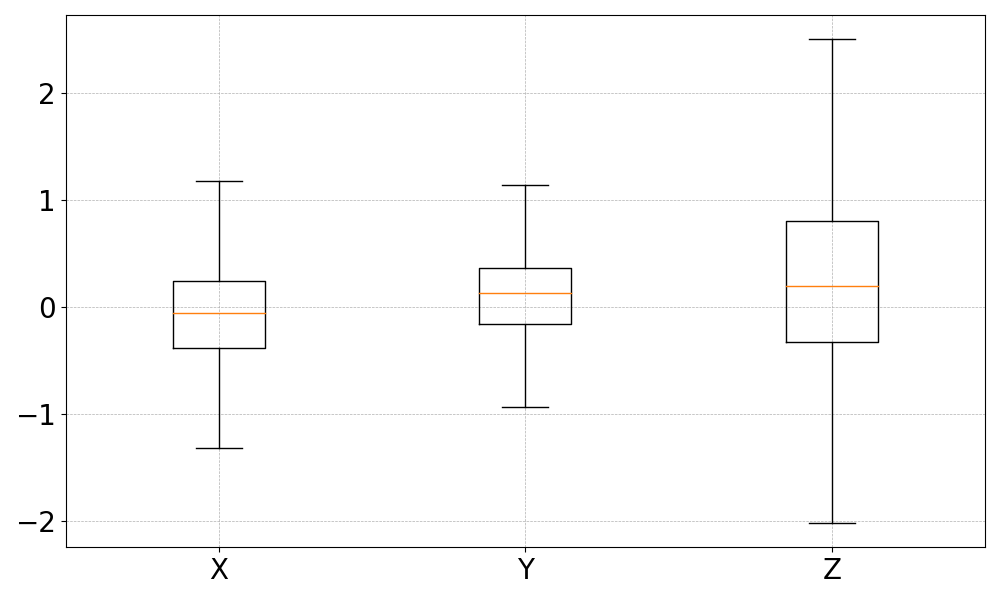}
    \caption{Translation errors (cm)}
    \label{fig:sub1}
  \end{subfigure}%
  \begin{subfigure}{.5\textwidth}
    \centering
    \includegraphics[width=\linewidth]{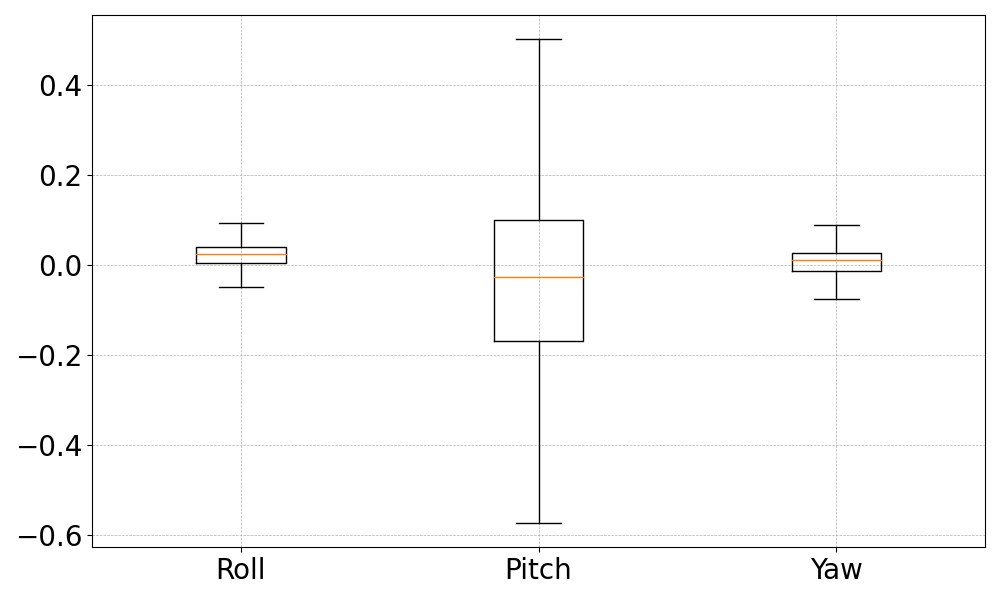}
    \caption{Rotation errors (degrees)}
    \label{fig:sub2}
  \end{subfigure}
  \caption{Box plots of translation and rotation errors on the test set of DSEC~\cite{gehrig2021dsec}.}
  \label{fig:box_plot}
\end{figure*}

\begin{figure*}
    \centering
\includegraphics[width=\textwidth]{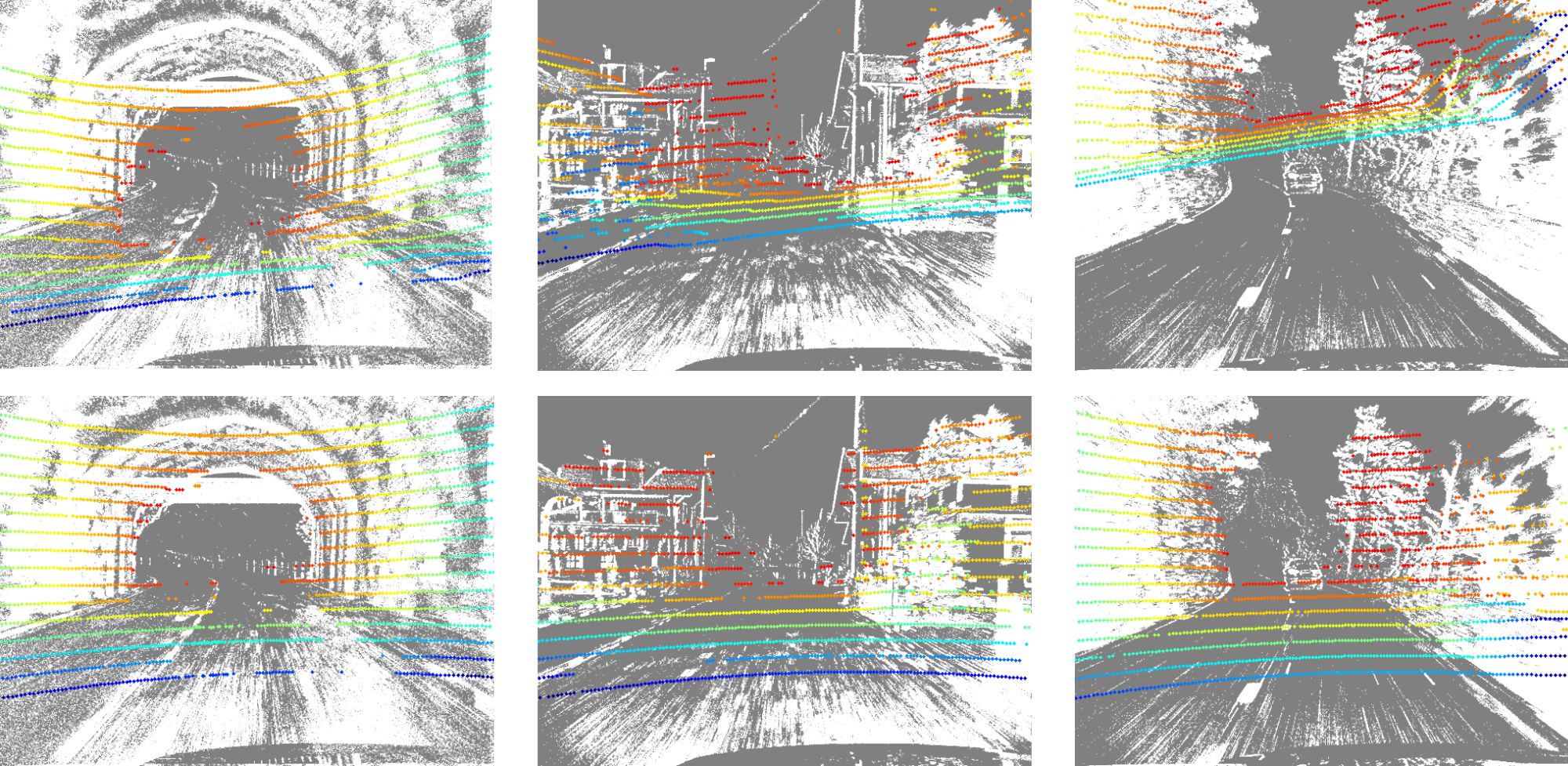}
    \caption{Qualitative results on DSEC~\cite{gehrig2021dsec}, showing three examples of recalibration in diverse environments. Images show the \lidar{} pointclouds projected on the event frame. The top lane presents random decalibrations applied to the setup, while the bottom lane presents the correction proposed by \muli{}.}
    \label{fig:qualitative}
\end{figure*}

\paragraph{General Results Analysis:}
As demonstrated by the results in~\Cref{tab:calibration_comparison}, \muli{} achieves superior calibration accuracy, reducing the translation error to an average of $0.81cm$ and rotation error to $0.1^{\circ}$. These results are illustrated qualitatively in~\Cref{fig:qualitative} and detailed in box plots in~\Cref{fig:box_plot}. Distinctively, \muli{} achieves these results while being, to our knowledge, the first online, targetless calibration method for this sensor setup. It bridges a significant gap in real-time operational needs while surpassing existing offline, target-dependent methods, such as~\cite{jiaoLCECalibAutomaticLiDARFrame2023b}. 
Finally, \muli{} being deep learning-based, it manages to reach this accuracy in an execution time inferior to $0.1s$ on a GPU, while an offline method like~\cite{ta2023l2e} takes about $134s$ with its fastest optimizer.

\begin{table}
\centering
\resizebox{\columnwidth}{!}{%
\begin{tabular}{lcc}
\toprule
\textbf{Location} & \textbf{Translation Error (cm)} & \textbf{Rotation Error (deg)} \\
\midrule
Interlaken & 1.07 & 0.12 \\
Thun & 0.59 & 0.08 \\
Zurich City & 0.40 & 0.08 \\
\bottomrule
\end{tabular}%
}
\caption{Evaluation of the mean absolute error of \muli{} on the location subsets of DSEC~\cite{gehrig2021dsec}.}
\label{tab:calibration_scenes}
\end{table}

\paragraph{Box Plots Analysis:}
Interestingly, it can be noticed in ~\Cref{fig:box_plot} that results on translation axis Z and rotation axis Pitch tend to be less regular. This was also found in works focused on RGB-\lidar{} calibration such as~\cite{lvLCCNetLiDARCamera2021}, and was thus expected. It can be explained by the physical nature of these axes that align with the vertical dimension, in which the \lidar{} points density is much lower (the vertical resolution of the VLP-16 \lidar{} is only $2^{\circ}$, while its horizontal resolution is between $0.1^{\circ}$ and $0.4^{\circ}$). This low vertical resolution of the VLP-16 is mostly due to it having only 16 \lidar{} rings, compared to 64 for the Velodyne HDL-64E used in the KITTI~\cite{geiger2013vision} dataset, which was most commonly used to evaluate RGB-\lidar{} calibration methods~\cite{cocheteux2023unical,iyerCalibNetGeometricallySupervised2018,lvLCCNetLiDARCamera2021,schneiderRegNetMultimodalSensor2017,wuThisWaySensors2021, Cocheteux_2023_BMVC,jing2022dxq}.

\paragraph{Influence of the Environment:}
To further analyze the behavior of \muli{} on different types of scenes, we measured the average errors per location. The results are available in~\Cref{tab:calibration_scenes}, while the characteristics and number of sequences in these locations were reported in~\Cref{tab:dataset-summary}. We observe in~\Cref{tab:calibration_scenes} that the best results are obtained in Zurich City, while the least accurate results were for scenes captured in Interlaken. This was expected, and we can infer from it two possible explanations: first, 35 sequences from Zurich were included in the training set, compared to 5 for Interlaken; second, scenes in Interlaken happen to be mostly rural, and thus to have generally less available features, especially long vertical edges like the ones offered by buildings. However, \muli{} performed better on the Thun scenes than Interlaken scenes, despite having even less sequences (only 1 for training). This tends to confirm our second hypothesis, as Thun offers more of a suburban environment, with enough human-built structures to offer more linear features.
Another interesting fact is that while Interlaken and Thun scenes were all recorded by day, Zurich City sequences include night scenes, and still obtains the best accuracy, suggesting that \muli{} might adapt quite well to varying lighting conditions.
Overall, results are at least on par with the state of the art or better for all three location subsets. Qualitative results in~\Cref{fig:qualitative} show successful recalibrations in different environment: in a tunnel, in a suburban zone, and on a rural road.

\subsection{Ablation Study}
\label{sec:ablation}
To assess the impact of event data representation and accumulation time on the calibration accuracy, we conducted experiments on the DSEC~\cite{gehrig2021dsec} dataset.

\paragraph{Event Representations:} 
The three different event representations considered and detailed in~\Cref{sec:representations} (event frame, voxel grid, and time surface) were evaluated to determine the best performing one.

\paragraph{Accumulation Times:} For the event frame representation, accumulation times of $30ms$, $50ms$, and $80ms$ were tested. These intervals were selected to explore the trade-off between temporal resolution and the richness of accumulated event information, potentially affecting the calibration's accuracy and robustness.

\paragraph{Ablation Results:} Results reported in~\Cref{tab:ablation_study} indicate that the event frame representation, with an accumulation time of $50ms$, achieved the highest calibration accuracy. This suggests that a longer accumulation time might lead to higher noise levels, degrading the result. Conversely, shorter accumulation times, while offering fresher data, may not accumulate enough events to adequately represent the scene for effective calibration. The voxel grid and time surface representations, despite their more complex encoding of event data, did not yield improvements in calibration accuracy over the optimized event frame representation. 
These observations underscore the importance of the choice of event representation and accumulation period to optimize the results of our method.

\subsection{Discussion}
Results of our experiments in ~\Cref{tab:calibration_comparison} demonstrate that \muli{} can provide better accuracy than existing offline works, and this in diverse environments (as demonstrated in~\Cref{tab:calibration_scenes}) while being the first online method proposed for this sensor combination. As a comparison, deep learning-based methods for RGB-LiDAR sensor setups have initially reached MAE of $0.28^{\circ}$ and $6cm$~\cite{schneiderRegNetMultimodalSensor2017}, while more recent approaches reached $0.03^{\circ}$ and $0.36cm$\cite{lvLCCNetLiDARCamera2021}, showing there is probably still  potential for improving the accuracy offered by \muli{} (currently $0.1^{\circ}$ and $0.81cm$). \muli{} not only enhances operational convenience by eliminating the need for impractical calibration targets but also excels in dynamic environments where rapid recalibration is essential, thanks to its execution time of less than $0.1s$ offering on-the-fly recalibration capability. By ensuring immediate recalibration to maintain performance and safety, \muli{} can contribute to the robustness of autonomous navigation systems in real-world applications.
Finally, results from~\Cref{tab:ablation_study} suggest that our choice of the simple event frame for event representation delivers the best results while simplifying the implementation of our method.

\begin{table}
\centering
\begin{tabular}{lcc}
\toprule
\textbf{Configuration} & \begin{tabular}[c]{@{}c@{}}\textbf{Translation} \\ \textbf{Error (cm)}\end{tabular} & \begin{tabular}[c]{@{}c@{}}\textbf{Rotation} \\ \textbf{Error (deg)}\end{tabular} \\
\midrule
Event Frame (30 ms) & 1.01 & 0.12 \\
Event Frame (50 ms) & \textbf{0.81} & \textbf{0.10} \\
Event Frame (80 ms) & 0.85 & 0.11 \\
Voxel Grid (50 ms) & 0.88 & 0.11 \\
Time Surface (50 ms) & 1.17 & 0.23 \\
\bottomrule
\end{tabular}%
\caption{Results of ablation experiments on DSEC~\cite{gehrig2021dsec} to determine the influence of event representation on the final calibration result (average error).}
\label{tab:ablation_study}
\end{table}

%% file: sec/5_conclusion.tex
\section{Conclusion}

In this work, we introduced \muli{}, a pioneering framework that establishes the feasibility of online, targetless calibration between event cameras and \lidar{}. This innovation marks a significant departure from traditional, offline calibration methods, offering enhanced calibration accuracy and operational flexibility. The real-time capabilities of \muli{} not only pave the way for immediate sensor recalibration—a critical requirement for the dynamic environments encountered in autonomous driving—but also open up new avenues for adaptive sensor fusion in operational vehicles.

Looking ahead, we aim to further refine \muli{}'s robustness and precision, with a particular focus on monitoring and adapting to the temporal evolution of calibration parameters. Such enhancements will ensure that \muli{} continues to deliver accurate sensor alignment even as conditions change over time. Additionally, we are interested in expanding the applicability of our framework to incorporate a wider array of sensor types and configurations. This expansion will enable more comprehensive and nuanced perception capabilities, ultimately facilitating the development of more sophisticated autonomous systems.

As we move forward, our focus on refining \muli{} is aligned with the evolving demands of autonomous vehicle technology. By addressing the real-world challenges of sensor calibration and integration, \muli{} contributes to improving the safety, reliability, and performance of these systems. Our efforts to enhance sensor fusion and adaptability reflect a practical step towards achieving more robust and reliable autonomous driving capabilities.